\documentclass[10pt,twocolumn,letterpaper]{article}

\usepackage[pagenumbers]{cvpr} % To force page numbers, e.g. for an arXiv version
\usepackage[table]{xcolor}
\usepackage{colortbl}
\usepackage{multirow}
\usepackage{makecell}
\usepackage{algorithm}% http://ctan.org/pkg/algorithm
\usepackage{algpseudocode}% http://ctan.org/pkg/algorithmicx
\usepackage{enumitem}

\def\cvprPaperID{7058} % *** Enter the CVPR Paper ID here
\def\confName{CVPR}
\def\confYear{2023}
\usepackage{graphicx}
\usepackage{amsmath}
\usepackage{amssymb}
\usepackage{booktabs}
\usepackage{hyperref}

\hypersetup{%
  colorlinks=true,
  urlbordercolor={1 1 1},
  linkcolor={black},
  raiselinks=false
}

\definecolor{darkgreen}{rgb}{0.0, 0.42, 0.24}

\usepackage[capitalize]{cleveref}
\crefname{section}{Sec.}{Secs.}
\Crefname{section}{Section}{Sections}
\Crefname{table}{Table}{Tables}
\crefname{table}{Tab.}{Tabs.}
\begin{document}

\title{AligNeRF: High-Fidelity Neural Radiance Fields via Alignment-Aware Training}

\author{%
  Yifan Jiang$^{1}$\footnotemark,  Peter Hedman$^{2}$, Ben Mildenhall$^{2}$, Dejia Xu$^{1}$, Jonathan T. Barron$^{2}$,\\ Zhangyang Wang$^{1}$, Tianfan Xue$^{3}$\footnotemark\\
  $^{1}$University of Texas at Austin, 
  $^{2}$Google Research,
  $^{3}$The Chinese University of Hong Kong
  \\
%   \texttt{\tt\small \{tianfan,bwronski,bmild,barron\}@google.com,~\{yifanjiang97,atlaswang\}@utexas.edu} \\
  \vspace{-2.em}
%   \And
}

\twocolumn[{%
\renewcommand\twocolumn[1][]{#1}%
\maketitle

\begin{center}
    \centering
    \captionsetup{type=figure}
    \includegraphics[width=0.99\textwidth,height=5cm]{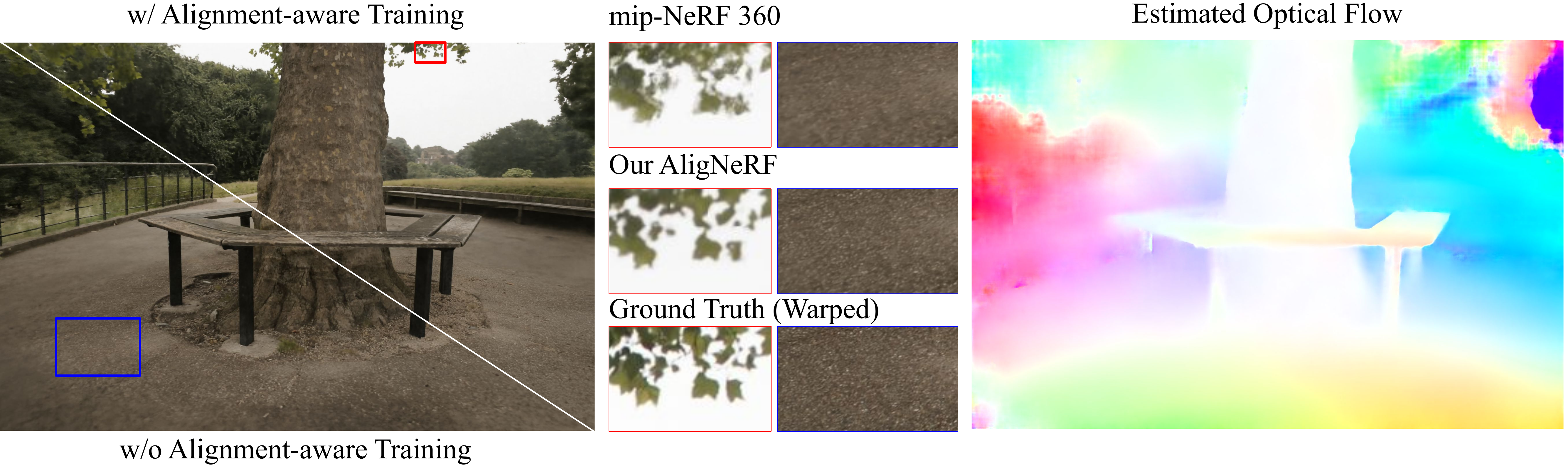}
    % \vspace{-1.em}
    \captionof{figure}{\textbf{Left:} Images rendered by mip-NeRF 360~\cite{mipnerf360} (w/o alignment-aware training) and our AligNeRF (w/ alignment-aware training), with the same amount of training time for both methods. \textbf{Middle:} Magnified cropped regions. Here, the ground truth has been aligned with the mip-NeRF 360 rendering, as described in Sec.~\ref{sec:analysis}. \textbf{Right:} Visualization of misalignment. This is the estimated optical flow between the mip-NeRF 360 rendering and the original ground truth images.}
\end{center}%
}]
\let\thefootnote\relax\footnote{* This work was performed while Yifan Jiang interned at Google.}
\let\thefootnote\relax\footnote{$\dagger$ This work was performed while Tianfan Xue worked at Google.}
\begin{abstract}
  Neural Radiance Fields (NeRFs) are a powerful representation for modeling a 3D scene as a continuous function.
  Though NeRF is able to render complex 3D scenes with view-dependent effects, few efforts have been devoted to exploring its limits in a high-resolution setting. Specifically, existing NeRF-based methods face several limitations when reconstructing high-resolution real scenes, including a very large number of parameters, misaligned input data, and overly smooth details.
  In this work, we conduct the first pilot study on training NeRF with high-resolution data and propose the corresponding solutions: 1) marrying the multilayer perceptron (MLP) with convolutional layers which can encode more neighborhood information while reducing the total number of parameters; 2) a novel training strategy to address misalignment caused by moving objects or small camera calibration errors; and 3) a high-frequency aware loss.
  Our approach is nearly free without introducing obvious training/testing costs, while experiments on different datasets demonstrate that it can recover more high-frequency details compared with the current state-of-the-art NeRF models. Project page: \url{https://yifanjiang.net/alignerf}.
\end{abstract}

\begin{figure*}[!t]
\centering
\includegraphics[width=0.99\textwidth]{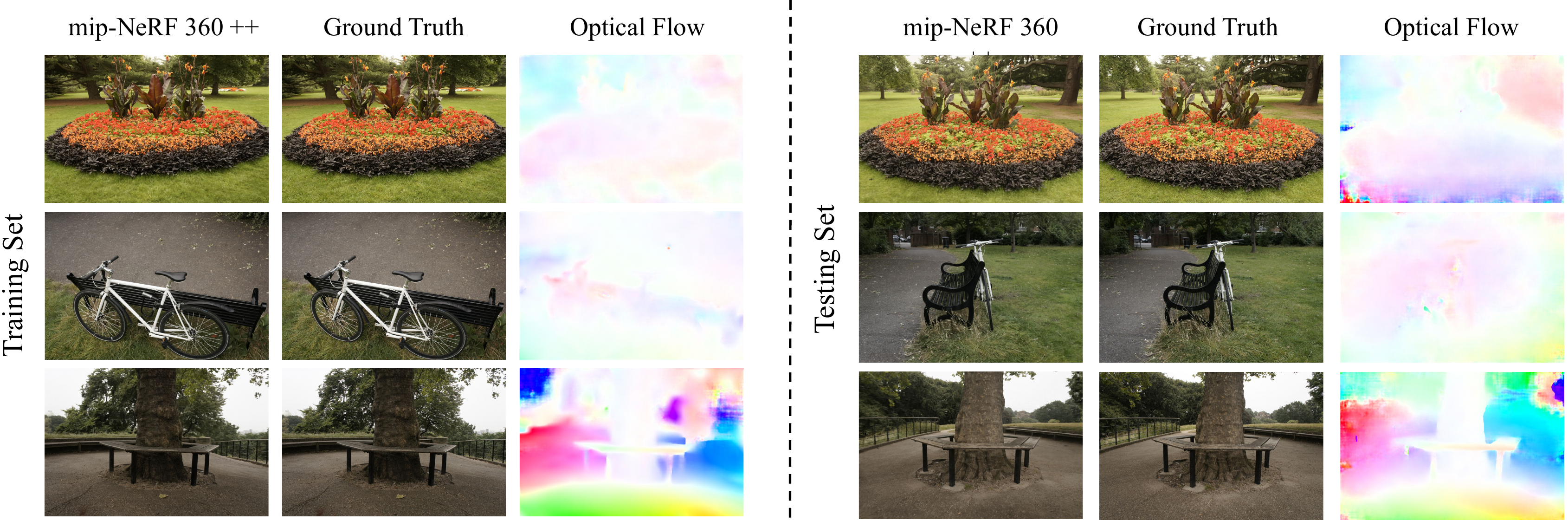}    
\caption{\textbf{Analysis of misalignment between rendered and ground truth images}. \textbf{mip-NeRF 360++}: Images rendered by a stronger mip-NeRF 360~\cite{mipnerf360} model ($16\times$ larger MLPs than the original). \textbf{Ground Truth}: The captured images used for training and testing. \textbf{Optical Flow}: Optical flow between the mip-NeRF 360++ and ground truth images, estimated by PWC-Net~\cite{sun2018pwc}. Significant misalignment is present in both training and test view renderings.}
\label{fig:flow}
\end{figure*}

\section{Introduction}
\label{sec:intro}

Neural Radiance Field (NeRF~\cite{nerf}) and its variants~\cite{mipnerf,mipnerf360,mvsnerf,dsnerf,martin2021nerf,liu2020neural}, have recently demonstrated impressive performance for learning geometric 3D representations from images. 
% In contrast to traditional voxel or mesh representations, NeRF~\cite{nerf} parameterizes the 3D scene by mapping positionally encoded input coordinates to various scene properties using a multilayer perceptron (MLP). 
The resulting high-quality scene representation enables an immersive novel view synthesis experience with complex geometry and view-dependent appearance.
Since the origin of NeRF, an enormous amount of work has been made to improve its quality and efficiency, enabling reconstruction from data captured ``in-the-wild''~\cite{martin2021nerf,kuang2022neroic} or a limited number of inputs~\cite{niemeyer2021regnerf,dsnerf,jain2021putting,roessle2021dense,xu2022sinnerf} and generalization across multiple scenes~\cite{pixelnerf,wang2021ibrnet,chen2021mvsnerf}.

However, relatively little attention has been paid to high-resolution reconstruction. mip-NeRF~\cite{mipnerf} addresses excessively blurred or aliased images when rendering at different resolutions, modelling ray samples with 3D conical frustums instead of infinitesimally small 3D points. mip-NeRF 360~\cite{mipnerf360} further extends this approach to unbounded scenes that contain more complex appearance and geometry. Nevertheless, the highest resolution data used in these two works is only $1280 \times 840$ pixels, which is still far away from the resolution of a standard HD monitor ($1920 \times 1080$), not to mention a modern smartphone camera ($4032 \times 3024$).

In this paper, we conduct the first pilot study of training neural radiance fields in the high-fidelity setting, using higher resolution images as input. This introduces several hurdles. \underline{First}, the major challenge of using high-resolution training images is that encoding all the high-frequency details requires significantly more parameters, which leads to a much longer training time and higher memory cost, sometimes even making the problem intractable~\cite{martin2021nerf,mipnerf360,kilonerf}.

\underline{Second}, to learn high-frequency details, NeRF requires accurate camera poses and motionless scenes during capture. However, in practice, camera poses recovered by Structure-from-Motion (SfM) algorithms inevitably contain pixel-level inaccuracies ~\cite{lindenberger2021pixsfm}. These inaccuracies are not noticeable when training on downsampled low-resolution images, but cause blurry results when training NeRF with higher-resolution inputs. Moreover, the captured scene may also contain unavoidable motion, like moving clouds and plants. This not only breaks the static-scene assumption but also decreases the accuracy of estimated camera poses. Due to both inaccurate camera poses and scene motion, NeRF's rendered output is often slightly misaligned from the ground truth image, as illustrated in Fig.~\ref{fig:flow}.
We investigate this phenomenon in Sec.~\ref{sec:analysis}, demonstrating that image quality can be significantly improved by iteratively training NeRF and re-aligning the input images with NeRF's estimated geometry.
The analysis shows that misalignment results in NeRF learning distorted textures, as it is trained to minimize the difference between rendered frames and ground truth images. Previous work mitigates this issue by jointly optimizing NeRF and camera poses~\cite{barf,nerf--,gnerf,garf}, but these methods cannot handle subtle object motion and often introduce non-trivial training overheads, as demonstrated in Sec~\ref{sec:simulation}.

To tackle these issues, we present AligNeRF, an alignment-aware training strategy that can better preserve high-frequency details. 
Our solution is two-fold: an approach to efficiently increase the representational power of NeRF, and an effective method to correct for misalignment. To efficiently train NeRF with high-resolution inputs, we marry convolutions with NeRF's MLPs, by sampling a chunk of rays in a local patch and applying ConvNets for post-processing. Although a related idea is discussed in NeRF-SR~\cite{wang2021nerf}, their setting is based on rendering test images at a higher resolution than the training set. Another line of work combines volumetric rendering with generative modeling~\cite{niemeyer2021giraffe,schwarz2020graf}, where ConvNets are mainly used for efficient upsampling and generative texture synthesis, rather than solving the inverse problem from many input images.
In contrast, our approach shows that the inductive prior from a small ConvNet improves NeRF's performance on high-resolution training data, without introducing significant computational costs.

In this new pipeline, we render image patches during training. This allows us to further tackle misalignments between the rendered patch and ground truth that may have been caused by minor pose errors or moving objects. First, we analyze how misalignment affects image quality by leveraging the estimated optical flow between rendered frames and their corresponding ground truth images. We discuss the limitations of previous misalignment-aware losses~\cite{zhang2019zoom,mechrez2018contextual}, and propose a novel alignment strategy tailored for our task. Moreover, our patch-based rendering strategy also enables patch-wise loss functions, beyond a simple mean squared error. That motivates us to design a new frequency-aware loss, which further improves the rendering quality with no overheads. As a result, AligNeRF largely outperforms the current best method for high-resolution 3D reconstruction tasks with few extra costs. 

To sum up, our contributions are as follows:
\begin{itemize}
 \item An analysis demonstrating the performance degradation caused by misalignment in high-resolution training data.
 \item A novel convolution-assisted architecture that improves the quality of rendered images with minor additional cost.
 \item A novel patch alignment loss that makes NeRF more robust to camera pose error and subtle object motion, togerther with a patch-based loss to improve high-frequency details.
\end{itemize}

\section{Preliminaries}

The vanilla NeRF~\cite{nerf} method takes hundreds of images with the corresponding camera poses as the training set, working on synthetically rendered objects or real-world forward-facing scenes. Later works extend NeRF to unconstrained photo collections~\cite{martin2021nerf}, reduce its training inputs to only sparsely sampled views~\cite{niemeyer2021regnerf,dsnerf,jain2021putting,roessle2021dense}, apply it to relighting tasks~\cite{zhang2021nerfactor,srinivasan2021nerv}, speed up training/inference time~\cite{mueller2022instant,hedman2021baking,sun2021direct,liu2020neural}, and generalize it to unseen scenes~\cite{pixelnerf,wang2021ibrnet,chen2021mvsnerf}.
% Another line of work addresses the training/inference speed of NeRF by adopting multi-resolution hash tables~\cite{mueller2022instant}, combining voxel grids with deep networks~\cite{hedman2021baking,sun2021direct,liu2020neural}, adopting ray termination techniques~\cite{piala2021terminerf}, or replacing the single large network with thousands of tiny MLPs~\cite{kilonerf}.
In contrast, we explore the problem of training on high-resolution input data in this work.

\subsection{NeRF}
NeRF takes a 3D location $\textbf{x} = (x, y, z)$ and 2D viewing direction $\textbf{d} = (\theta, \phi)$ as input and outputs an emitted color $\textbf{c} = (r, g, b)$ and volume density $\sigma$. This continuous 5D scene representation is approximated by an MLP network $F_{\Theta}: (\textbf{x}, \textbf{d}) \rightarrow (\textbf{c}, \sigma)$. 
% where the network weights $\Theta$ are optimized to map each input 5D coordinate to a corresponding color and volume density.
To compute the output color for a pixel, NeRF approximates the volume rendering integral using numerical quadrature~\cite{max1995optical}. For each camera ray $\textbf{r}(t) = \textbf{o} + t\textbf{d}$, where $\textbf{o}$ is the camera origin and $\textbf{d}$ is the ray direction, the expected color $\widehat{C}(\textbf{r})$ of $\textbf{r}(t)$ with near and far bounds $t_n$ and $t_f$ can be formulated as:
\begin{align} 
\widehat{C}(\textbf{r}) &= \sum_{k} w_k \textbf{c}_k \, , \\
\text{with } w_k &= T_k (1 - \exp(-\tau_{k}(t_{k+1} - t_k))) \label{eqn:weight} \\
\text{and } T_k &= \exp\left( -\sum_{k^{'}<k} \tau_{k^{'}}(t_{k^{'}+1} - t_{k^{'}}) \right),
\end{align}
where $\{t_k\}_{k=1}^K$ represents a set of sampled 3D points between near and far planes of the camera, and $\tau$ indicates the estimated volume density $\sigma$. Since neural networks are known to be biased to approximate lower frequency functions~\cite{rahaman2019spectral}, NeRF uses a positional encoding function $\gamma(\cdot)$ consisting of sinusoids at multiple frequencies to introduce high-frequency variations into the network input. 

% To efficiently sample the most important 3D points that contribute to the emitted color, NeRF simultaneously optimizes two individual MLPs, one coarse and one fine. The coarse NeRF is sampled uniformly along a ray and its predicted density is used to guide sampling of points used as input to the fine NeRF. Both MLPs are supervised by an RGB color loss on rendered ray colors.

\subsection{mip-NeRF 360}
To tackle the aliasing issue when the resolution of rendered views differs from training images, mip-NeRF~\cite{mipnerf} proposes to sample volumetric frustums from a cone surrounding the ray rather than sampling infinitesimally small 3D points. 
mip-NeRF 360~\cite{mipnerf360} extends this integrated positional encoding to unbounded real-world scenes by adopting a ``contraction'' function that warps arbitrarily far scene content into a bounded domain (similarly to NeRF++~\cite{nerfpp}).

To make training more efficient, mip-NeRF 360 also adopts an online distillation strategy. The coarse NeRF (also named ``Proposal MLP'') is no longer supervised by photometric reconstruction loss, but instead learns the distilled knowledge of structure from the fine NeRF (also named ``NeRF MLP''). 
% This is achieved by encouraging the color accumulation weights $w(k)$ (from Eqn.~\ref{eqn:weight}) emitted by the Proposal MLP and NeRF MLP to be consistent. 
By doing so, the parameter count of the Proposal MLP can be largely reduced, as it does not contribute to the final RGB color. Because the ray samples are better guided by the Proposal MLP, it is also possible to significantly reduce the number of queries to the final NeRF MLP. Thus a fixed computational budget can be reallocated to significantly enlarge this NeRF MLP, improving the rendered image quality for the same total cost as mip-NeRF.

\section{Method}
\label{sec:method}
In this section, we introduce AligNeRF, an alignment-aware training strategy to address the obstacles discussed in Sec.~\ref{sec:intro} and analyzed in Sec.~\ref{sec:analysis}. AligNeRF is an easy-to-plug-in component for any NeRF-like models, including both point-sampling approaches and frustum-based approaches. 
AligNeRF uses staged training: starting with an initial ``normal" pre-training stage, followed by an alignment-aware fine-tuning stage.
We choose mip-NeRF 360 as our baseline, as it is the state-of-the-art NeRF method for complex unbounded real-world scenes. Next, we introduce our convolution-augmented architecture, then present our misalignment-aware training procedure and high-frequency loss.

\subsection{Marrying Coordinate-Based Representations with Convolutions}
\label{sec:method_conv}
Inspired by the recent success of encoding inductive priors in vision transformers~\cite{dai2021coatnet}, our first step is to explore how to effectively encode local inductive priors for coordinate-based representations such as NeRF. 
Recall that NeRF-like models generally construct a coordinate-to-value mapping function and randomly sample a batch of rays to optimize its parameters, which prevents us from performing any patch-based processing. Thus, our first modification is to switch from random sampling to patch-based sampling, with respect to camera rays (we use $32\times 32$ patches in our experiments).

This patch-based sampling strategy allows us to gather a small local image region during each iteration and thus make use of 2D local neighborhood information when rendering each pixel. 
To begin with, we change the number of output channels of the last layer in MLPs from $3$ to a larger $N$, and apply numerical integration along the ray in this feature space rather than in RGB space. This helps gather richer representation in each camera ray. Next, we add a simple 3-layer convolutional network with ReLU activations and $3\times 3$ kernels, following the volumetric rendering function. We adopt ``reflectance'' padding for the inputs of each convolution and remove the padding values in the outputs, to prevent checkerboard artifacts.
At the end of this network, we use a feed-forward perceptron layer to convert the representation from feature space to RGB space. As a result, the rendering process for each pixel does not only rely on the individual ray/cone along that direction, but also depends on its neighboring regions, which help produces better texture detail. Because our CNN is very shallow and does not perform any upsampling, we do not observe any resulting multiview inconsistency in the image outputs.

\begin{figure*}[!t]
\centering
\includegraphics[width=0.99\textwidth]{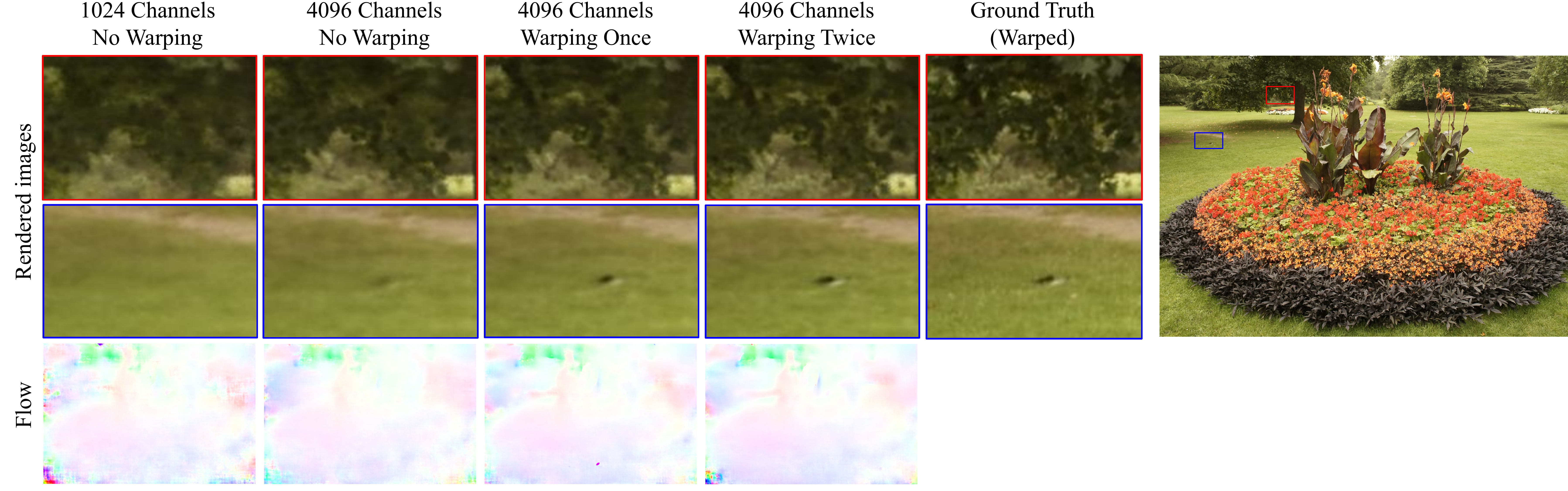}    
\caption{\textbf{Misalignment prevents recovery of high-frequency detail.} We randomly select an example test set image. \textbf{1024 channels, No Warping}: We train mip-NeRF 360~\cite{mipnerf360} using the original training set with default parameters (1024 channels for the ``NeRF MLP''). \textbf{4096 channels, No Warping}: We scale mip-NeRF 360 up with $4 \times$ more channels in the ``NeRF MLP''.
\textbf{4096 channels, Warping Once}: We train the 4096 channel model using the iterative alignment strategy described in Section~\ref{sec:analysis}, which uses aligned ground truth images. % We apply an iterative refinement data engine to mip-NeRF 360 with bigger MLPs, by using warped ground truth images as training data. See Section~\ref{sec:analysis} for details.
% Yifan: here we dont finetune the mipnerf360, but instead train it from scratch.tianfna: o.k
\textbf{4096 channels, Warping Twice}: Similar to ``4096 channels, Warping Once'', but with two iterations for better alignment. The bottom row shows the flow between rendered frames and ground truth. Note that NeRF can recover much more high-frequency detail with aligned training data (columns 3 and 4).
% We conduct an iterative refinement data engine again with another new iteration, adopting the rendered images from ``4096 channels Warping Once'' to estimate a set of new optical flow and retrain mip-NeRF 360 using new warped data. The first and second row of each example show the zoom-in box and the third row shows the optical flow estimated between current rendered results and original ground truth images.
% We pack the new training images rendered by mip-NeRF 360 and original ground truth images together to predict the optical flow between each pair, and further adopting these optical flows to generate new training data by warping the ground truth images. This simple method help mitigate the misalignment issue and significantly improve mip-NeRF 360.
}
\label{fig:flow_iteration}
\end{figure*}

\subsection{Alignment-Aware Loss}
Recall that NeRF models a 3D scene using a rendering function $F_{\Theta}: (\textbf{x}, \textbf{d}) \rightarrow (\textbf{c}, \sigma)$ mapping the coordinates of 3D points to the properties of the scenes.
% Under this framework, the precise camera origins become the crucial prerequisite of training a strong NeRF model.
Under this framework, the accuracy of camera poses is crucial for NeRF training, otherwise, rays observing the same 3D point from different viewpoints may not converge to the same location in space. The vanilla NeRF~\cite{nerf} solves this problem by capturing images over a very short timespan (to prevent scene motion and lighting changes) and adopts COLMAP~\cite{schoenberger2016sfm} to calculate camera parameters. This data preparation pipeline is mostly reliable except 1) There is a gap between the ground truth camera poses and the camera poses from COLMAP, as has been pointed out by previous works~\cite{iheaturu2020assessment,raoult2017reliable}; and 2) It is usually hard to avoid images with swaying plants and other nonrigid objects in uncontrolled outdoor scenes, which further hurts the performance of COLMAP. In the high-resolution reconstruction setting, the misalignment issue caused by camera poses and moving objects can be further amplified, as pixel-space misalignment scales linearly with resolution. We explore how much this misalignment make affect the quality of rendered images in Sec.~\ref{sec:analysis}. To address this issue, we propose an alignment-aware training strategy that can be adopted to refine the quality of rendered images.

Despite the distorted textures, we observe that NeRF still learns the rough structures from the misaligned images, as shown in the second column of Fig.~\ref{fig:flow_iteration}. 
% Taking advantage of this, we propose to replace the point-to-point matching strategy with patch-based matching.
Taking advantage of this, we proposed a loss between aligned ground truth and rendered patches. Let $G$ denote the ground truth patch and $R$ denote the rendered patch. We sample a larger size of ground truth patch $G$ during each iteration and search over every possible subpatch $G_i$ for the best match with the smaller rendered patch $R$. Since NeRF may render a very blurry patch $R$ that seems equally well-aligned with many possible patches $G_i$ from the ground truth set, we additionally set a regularization term based on Euclidean distance as a penalty for this search space. The final loss function is
\begin{align}
    \mathcal{L}_{PM}(G, R) = \min_{i=1,\dots,N}(D(G_i, R) + \lambda \cdot D_{coord}(G_i, R))
\end{align}
where $D(G_i, R) = \left \lVert G_i - R \right\rVert_2^2$ and $D_{coord}(G_i, R) = \sqrt{\Delta x^2 + \Delta y^2}$, and $\Delta x$/$\Delta y$ are the horizontal/vertical offset between $G_i$ and $R$. We adopt $\lambda$ to control the regularization term and empirically set it to $0.01$. In all of our experiments, we sample $48 \times 48$ patches for $G$ and render $32 \times 32$ patches for $R$ within each iteration.

Similar losses are also used in image super-resolution~\cite{zhang2019zoom} or style transfer~\cite{mechrez2018contextual}. However, those losses are defined on single pixels, while our alignment-aware loss is defined on patches, where alignment vectors $(\Delta x, \Delta y)$ can be more robustly estimated. 
% \tianfan{Yifan: please take a look of this paragraph.}
% Yifan: LGTM
% To verify this phenomenon, we start by conducting a toy experiment to measure how much this misalignment affects the performance of NeRF-like models.

% \renewcommand{\algorithmicrequire}{\textbf{Input:}}
% \renewcommand{\algorithmicensure}{\textbf{Output:}}
% \begin{algorithm}
%   \caption{Euclid’s algorithm}\label{euclid}
%   \begin{algorithmic}[1]
%     \Require rendered patch $p$, ground truth patch $\hat{p}$, step size $\lambda$, weight $w$, distance metric $D(\cdot)$
%     % \Procedure{Euclid}{$a,b$}\Comment{The g.c.d. of a and b}
%       \State $r\gets a\bmod b$
%       \While{$r\not=0$}\Comment{We have the answer if r is 0}
%         \State $a\gets b$
%         \State $b\gets r$
%         \State $r\gets a\bmod b$
%       \EndWhile\label{euclidendwhile}
%       \For{\texttt{<some condition>}}
%         \State \texttt{<do stuff>}
%       \EndFor
%       \State \textbf{return} $b$\Comment{The gcd is b}
%     % \EndProcedure
%   \end{algorithmic}
% \end{algorithm}
% \subsection{High-frequency aware Loss}
\subsection{High-frequency aware Loss}
\label{sec:mehtod_percetual_loss}
Mean squared error (MSE) loss is commonly used to supervise NeRF training, but it is well-known in the image processing literature that MSE often leads to blurry output images~\cite{johnson2016perceptual,ledig2017photo}. Given our patch sampling strategy, we can adopt a perceptual loss, which better preserves high-frequency details.
% Since we modify the sampling strategy in the training process of NeRF from individual camera rays to patch-wise rays, we are allowed to apply a more patch-based loss function to guide the optimization of volumetric rendering, rather than a simple mean square error (MSE) loss.
% Motivated by the success of super-resolution tasks, we start by applying the perceptual loss~\cite{johnson2016perceptual} built upon the feature space of a pretrained VGG network~\cite{simonyan2014very}, in addition to the MSE loss.
We first attempted to use L2 loss of a pretrained VGG features~\cite{simonyan2014very}. However, similar to other image restoration tasks~\cite{ledig2017photo}, we found that perceptual loss produced more high-frequency details but sometimes distorted the actual texture of the object.
% (in practice, it improves LPIPS~\cite{zhang2018unreasonable} score but degrades PSNR/SSIM scores.), which works against our original purpose. 
Instead, we modify the original perceptual loss proposed by Johnson et al.~\cite{johnson2016perceptual}, by only using the output of the first block before max-pooling: % of a very shadow layer of the pre-trained VGG network, as:
\begin{align}
\mathcal{L}_{\mathrm{HF}}(G_i, R) = \frac{1}{CWH} \left\lVert F(G_i) - F(R) \right\rVert_2^2,
\end{align}
where $G_i$ denotes the ground truth patch after alignment and $R$ denotes the rendered patch, $F$ denotes the first block of a pre-trained VGG-19 model, and $C$, $W$, and $H$ are the dimensions of the extracted feature maps. With this change, the proposed loss improves the high-frequency details while still preserving real textures. % Since we only adopt several convolutional layers in the first block of the VGG network, this ConvBlock may serve as the high-dimensional version of the traditional Sobel operator or Prewitt operator, which supervises the training process in a higher-order space, in addition to the RGB space.

% This concludes our discussion of the main techniques we are using to construct AligNeRF. 
In summary, the major difference between AligNeRF and previous work is switching from per-pixel MSE loss to a combination of patch-based MSE loss (accounting for misalignment) and a shallow VGG feature-space loss to improve high-frequency detail:
\begin{equation}
    \mathcal{L}_{MSE} \longrightarrow \mathcal{L}_{PM} + w\cdot\mathcal{L}_{HF},
\end{equation} where $w$ is empirically set to be $0.05$. To facilitate comparisons and demonstrate the use of AligNeRF as a simple plug-and-play modification, other regularization losses from mip-NeRF 360 are kept the same by default.

\section{Experiments}
\subsection{Dataset}
The main testbed of our experiments are the outdoor scenes from mip-NeRF 360~\cite{mipnerf360}, as this is the most challenging benchmark with unbounded real-world scenes, complex texture details, and subtle scene motion. \cite{mipnerf360} captured 5 outdoor scenes, including ``\textit{bicycle}'', ``\textit{flowers}'', ``\textit{treehill}'', ``\textit{garden}'', and ``\textit{stump}''. However, these captured images are resized to be $4\times$ smaller, resulting in a resolution of approximately $1280 \times 840$. To better investigate misalignment we also perform experiments on a higher-resolution version ($2560 \times 1680$) of these scenes. Finally, we also generate new better-aligned test set to avoid drawing biased conclusions from misaligned data, as described in Sec.~\ref{sec:analysis}.
% Furthermore, we observe the severe misalignment issue occurs on both training and testing set of high-resolution benchmarks, as described in Fig.~\ref{fig:flow}.
% Since conducting evaluation on these misaligned images may introduce biased conclusion, we re-generate a new testing set where the misalignment can be mitigated, as described in Sec.~\ref{sec:analysis}.
\begin{figure}[!t]
    \centering
    \includegraphics[width=0.48\textwidth]{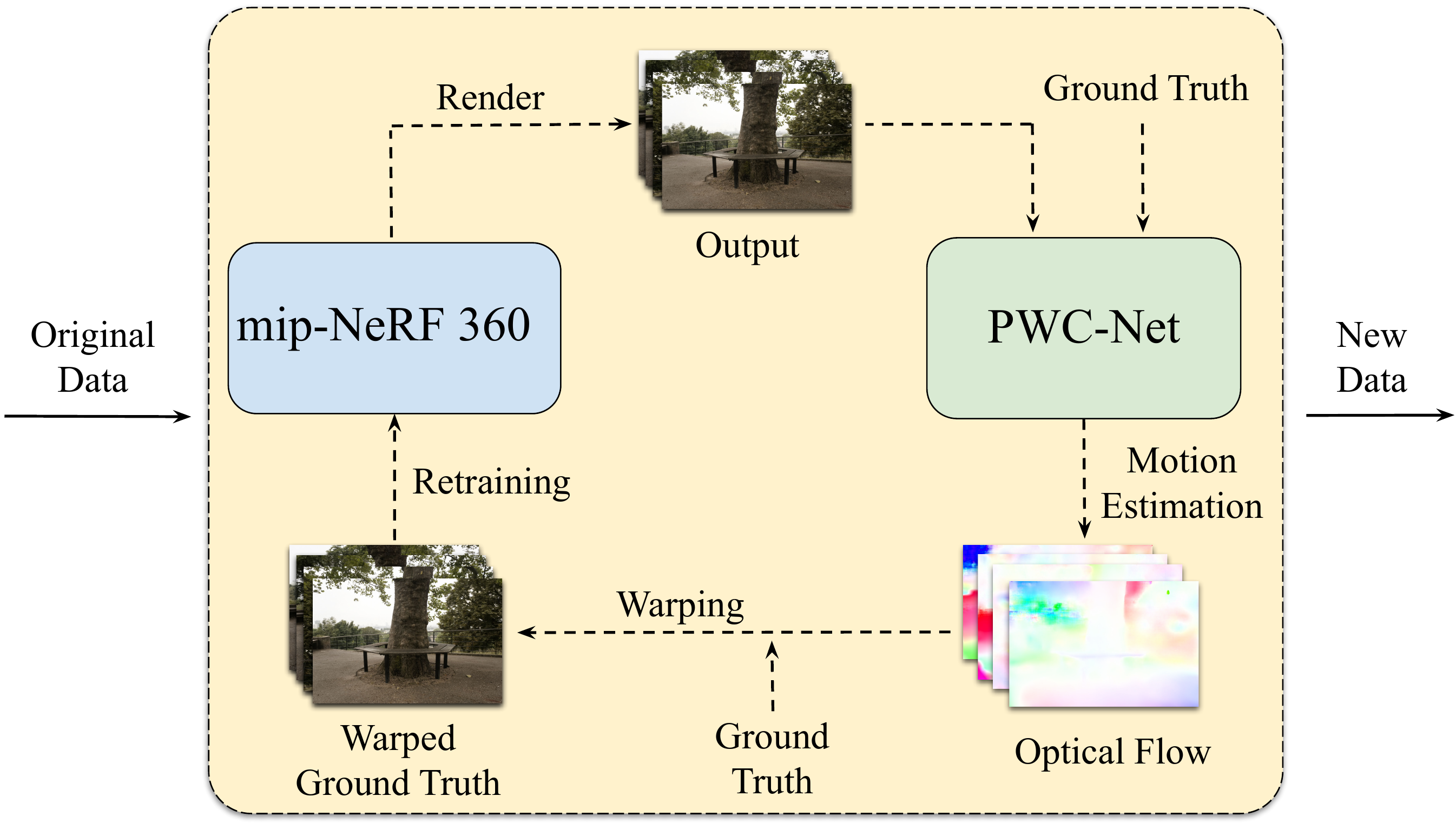}    
\caption{\textbf{Illustration of one iteration in our iterative alignment strategy}. Note that we were unable to use RAFT~\cite{teed2020raft}, as it runs out of memory on our high-resolution images. \textbf{This strategy only serves for analysis and does not represent the proposed algorithm in this work.}}
\label{fig:iterative_training}
\end{figure}

\subsection{Training Details}
\label{sec:training}

\paragraph{Standard Training Setting.} Baseline algorithms are optimized following the standard training procedure of mip-NeRF 360: a batch size of $16384$, an Adam~\cite{kingma2014adam} optimizer with $\beta1$ = 0.9 and $\beta2$ = 0.999, and gradient clipping to a norm of 1e-3.
The initial and final learning rates are set to be $2 \times 10^{-3}$ and $2 \times 10^{-5}$, with an annealed log-linear decay strategy. The first 512 iterations are used for a learning rate warm-up phase.

\paragraph{Alignment-aware Training Setting.} We split the training process of AligNeRF into two stages: the pre-training stage (MLPs only) and the fine-tuning stage (MLPs together with ConvNets). In the first stage, we generally follow the standard training procedure but with $\textbf{60\%}$ as many iterations. This is followed by another stage using the alignment-aware training strategy described in Sections~\ref{sec:method_conv}-\ref{sec:mehtod_percetual_loss}, where we sample a $32\times32$ patch in each batch instead of individual pixels and set the batch size to $32$ patches. Since the total number of rays is $2\times$ larger than the pre-training stage, we only use $\textbf{20\%}$ as many iterations for the second stage to equalize the total number of rays seen during training. This makes the total training cost of both stages approximately equal to the standard training. Moreover, we report benchmarks for two settings: the standard 250k iterations used in mip-NeRF 360~\cite{mipnerf360} and a $4\times$ longer version trained for 1000k steps. This is because using $2\times$ larger resolution training images requires $4\times$ more iterations for the same number of total epochs. All experiments are conducted on a TPU v2 accelerator with 32 cores.

\begin{table}[!t]
  \caption{Analysis of misalignment. All results are reported on the ``\textit{flowers}'' scene ($2560 \times 1680$) in the ``outdoor'' dataset~\cite{mipnerf360}. The ``warping'' column indicates whether we use the re-generated data and how many iterations it takes. $c$ represents the channels of the ``NeRF MLP'', which contributes to the RGB color. Training time contains both the NeRF training time and data generation time. We report metrics on both the \underline{standard} and \underline{warped} test sets respectively, split by ``/''.}
    \label{tab:warp}
    \resizebox{0.99\linewidth}{!}{
    \begin{tabular}{l|c|ccc|c|c}
    \toprule
    Methods & Warping&PSNR $\uparrow$&SSIM $\uparrow$&LPIPS $\downarrow$ & Time & \#Params \\ 
    \hline
    \multirow{3}{*}{\makecell[c]{mip-NeRF 360 \\ ($c=4096$)}} & None&\cellcolor{yellow!25}21.23/22.06 &\cellcolor{yellow!25}0.560/0.613 &\cellcolor{yellow!25}0.384/0.365 & 27.52 & \textbf{139.2M}\\
    & Once &\cellcolor{orange!25}21.33/22.46&\cellcolor{orange!25}0.570/0.637 &\cellcolor{orange!25}0.368/0.346 & 90.04& \textbf{139.2M}\\
    & Twice &\cellcolor{red!25}21.33/22.68&\cellcolor{red!25}0.581/0.661&\cellcolor{red!25}0.344/0.321& \textbf{153.1}& \textbf{139.2M}\\
    \hline
    \hline
    \multirow{3}{*}{\makecell[c]{mip-NeRF 360 \\ ($c=1024$)}} & None& 20.83/21.63 & 0.516/0.566 & 0.436/0.419 & 6.88 & 9.9M\\
    & Once &\cellcolor{yellow!25}20.89/21.85&\cellcolor{yellow!25}0.523/0.580 &\cellcolor{yellow!25}0.427/0.409 & 23.24& 9.9M\\
    & Twice &\cellcolor{orange!25}20.80/21.86&\cellcolor{red!25}0.519/0.582&\cellcolor{orange!25}0.426/0.408& \textbf{39.36}& 9.9M\\
    \hline
    Ours ($c=1024$)& None &\cellcolor{red!25}20.89/21.86 &\cellcolor{orange!25}0.521/0.580&\cellcolor{red!25}0.425/0.380 & 7.12& 10.4M \\
    \bottomrule
\end{tabular}
}
\end{table}

\begin{table*}[!t]
  \caption{Quantitative comparison between ours and state-of-the-art methods on the ``outdoor'' dataset~\cite{mipnerf360} at high-resolution ($2560 \times 1680$).}
  \label{tab:high_res}
  \centering
  \resizebox{0.8\linewidth}{!}{
  \begin{tabular}{l|c|ccc|ccc|c|c}
    \toprule
    \multirow{2}{*}{Methods} & \multirow{2}{*}{Iterations} & \multicolumn{3}{c|}{Standard Test Set} & \multicolumn{3}{c|}{Warped Test Set} & \multirow{2}{*}{Time (hrs)} & \multirow{2}{*}{\#Params} \\ 
    \cline{3-8} 
    & & PSNR $\uparrow$ & SSIM $\uparrow$ & LPIPS $\downarrow$ & PSNR $\uparrow$ & SSIM $\uparrow$ & LPIPS $\downarrow$ & & \\
    \hline
    NeRF~\cite{nerf} & $1\times$ & 21.44 & 0.474 & 0.665 &- &- & -& 4.16 & 1.5M\\
    mip-NeRF~\cite{mipnerf}& $1\times$ & 21.36 & 0.484 & 0.553 &- & -&- & 3.17 & 0.7M\\ 
    mip-NeRF~\cite{mipnerf} w/ bigger MLPs& $1\times$ &\cellcolor{yellow!25}21.90 &\cellcolor{yellow!25}0.566 &\cellcolor{yellow!25}0.447 &\cellcolor{yellow!25}22.44 &\cellcolor{yellow!25}0.605&\cellcolor{yellow!25}0.433 & 22.71 & 9.0M\\ 
    mip-NeRF 360~\cite{mipnerf360}& $1\times$ &\cellcolor{orange!25}23.71&\cellcolor{orange!25}0.644&\cellcolor{orange!25}0.368&\cellcolor{orange!25}24.58&\cellcolor{orange!25}0.693&\cellcolor{orange!25}0.349&  6.88 &  9.9M\\
    Ours& $1\times$ &\cellcolor{red!25}23.84&\cellcolor{red!25}0.649&\cellcolor{red!25}0.365& \cellcolor{red!25}24.77&\cellcolor{red!25}0.70&\cellcolor{red!25}0.340& 7.12 & 10.4M\\
    \hline
    NeRF~\cite{nerf}& $4\times$ & 21.60 & 0.483 & 0.631 & -& -& -& 16.64 & 1.5M\\
    mip-NeRF~\cite{mipnerf}& $4\times$ &\cellcolor{yellow!25}21.64 &\cellcolor{yellow!25}0.511 &\cellcolor{yellow!25}0.523 & -&- &- & 12.68 & 0.7M\\
    mip-NeRF 360~\cite{mipnerf360}& $4\times$ &\cellcolor{orange!25}23.88&\cellcolor{orange!25}0.665&\cellcolor{orange!25}0.339&\cellcolor{orange!25}24.83&\cellcolor{orange!25}0.718&\cellcolor{orange!25}0.320 &27.56 & 9.9M\\
    % (*) MipNeRF-360~\cite{mipnerf360} w/ bigger MLP & & & & & & & 108.41 & 55.69\\
    
    Ours& $4\times$ &\cellcolor{red!25}24.16&\cellcolor{red!25}0.678&\cellcolor{red!25}0.327&\cellcolor{red!25}25.22&\cellcolor{red!25}0.734&\cellcolor{red!25}0.299& 28.48 & 10.4M\\
  \bottomrule
\end{tabular}
}
\end{table*}

\subsection{Misalignment Analysis}
\label{sec:analysis}
We begin by analyzing the causes of quality degradation when scaling NeRF up to higher resolution. In particular, we show how training image misalignment affect the quality of images rendered by NeRF. To illustrate this, we perform an ablation study where we correct for misalignment using motion estimation techniques. Inspired by~\cite{Zhou2014}, we use motion estimation to align the training views with the geometry reconstructed by NeRF.
% A straightforward way to answer this question is to fix this misalignment issue and see how much it helps improve NeRF, using motion estimation techniques. Therefore, we try to re-generate a set of aligned training views, using motion estimation techniques.

\subsubsection{Re-generating Training/Testing Data with Iterative Alignment}
\label{sec:iterative}
Here we correct for misalignment in the dataset by using optical flow to align the input images with the geometry estimated by NeRF.
To achieve this, we use a high-quality but expensive motion estimator (PWC-Net~\cite{sun2018pwc}) to calculate the optical flow between images rendered by NeRF and their corresponding ground truth views. 
However, we observe that our case partially violates the assumption of general optical flow estimators, which usually expect two sharp images as paired inputs. In practice though, the rendered images produced by NeRF are mostly blurry due to misalignment. As shown in the second column in Fig.~\ref{fig:flow_iteration}, the current best NeRF variant (mip-NeRF 360~\cite{mipnerf360}) fails to generate sharp details, which hurts the optical flow result estimated by PWC-Net. 

% To fix this problem, we build a stronger mip-NeRF 360 stronger and produce sharper rendered views, by scaling its parameters to $16\times$ larger. Comparing with vanilla mip-NeRF 360, this stronger NeRF slightly improves the visual quality and refine some parts of the estimated optical flow, yet still can not reach a satisfactory quality.
To address this issue, we propose an iterative alignment strategy, shown in Fig.~\ref{fig:iterative_training}. That is, at every alignment iteration we:
1) Train mip-NeRF 360 using the original ground truth images, and render output images for each training view. 2) Next, we calculate the flow from the rendered views to ground truth views using PWC-Net. Although the flow field might contain some minor inaccuracies due to the blurry NeRF images, PWC-Net can generally produce reasonable results based on global structures and shapes. 3) Finally, we warp the ground truth images using these estimated optical flows, building a new training set which is better aligned with the geometry estimated by NeRF.
% 4) In the last step, we adopt this new training set to retrain the mip-NeRF 360 model, and go back to step 2) to start a new iteration. 
%
At each alignment iteration, mip-NeRF 360 can leverage better aligned  data to produce sharper images. These sharper images then further improve the accuracy of estimated optical flow. And more accurate optical flow enables us to generate better aligned training images for the next round of training. We observe that it generally takes 2-3 iterations to reach the best quality. 

Using this alignment strategy, we are also to also generate a new set of better aligned test images, which helps us avoid drawing biased conclusions from the original misaligned test set. Note that we use the same aligned test set for all methods, which we generated using the highest quality mip-NeRF 360 model with 4096 channels. In the following experiments we report results on both two test sets. 

\begin{figure*}[!t]
\centering
\includegraphics[width=0.99\textwidth]{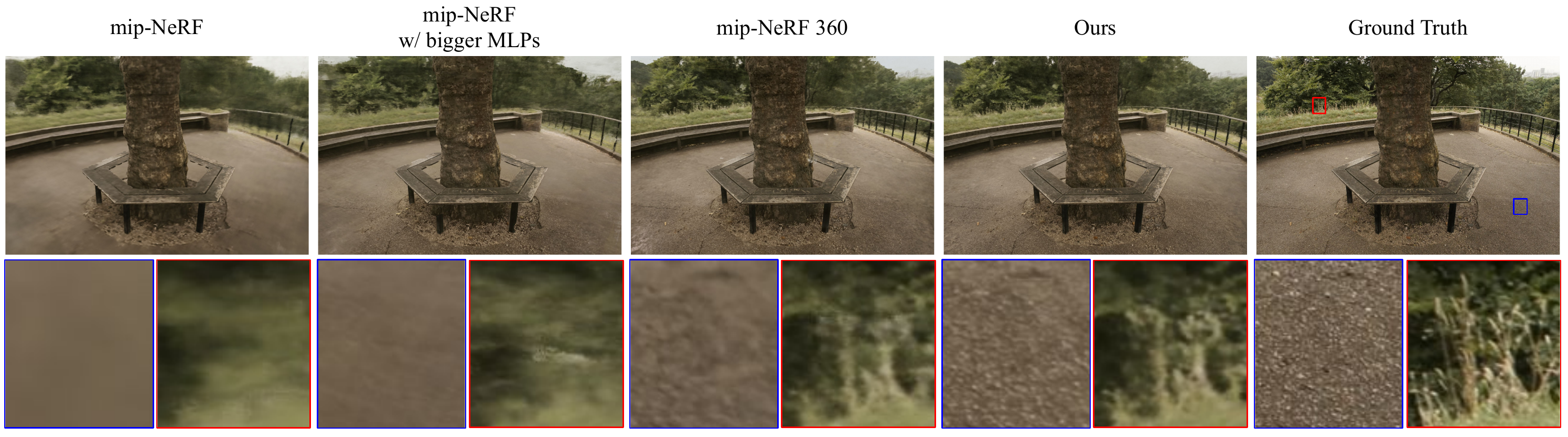}    
\caption{Qualitative comparison on a high-resolution version ($2560 \times 1680$) of the ``\textit{treehill}'' scene.
}
\label{fig:comparison}
\end{figure*}

\subsubsection{Qualitative Analysis of Misalignment Issue} 
In Fig.~\ref{fig:flow_iteration} we compare the intermediate visual examples from our iterative alignment strategy. First, we train a mip-NeRF 360 model with default parameters (1024 channels). This results in blurry images and the estimated optical flow contains artifacts in the distorted regions (first column). Next, we increase the mip-NeRF 360 parameters by $4\times$, which only marginally improves the visual quality of the results. We also apply our iterative alignment strategy to improve the results of this stronger model. By comparing the third and fourth columns, we see that the model trained with re-generated data recovers much sharper details compared to the ones trained on misaligned data (first two columns). This observation implies that the current best NeRF models are strongly affected by misaligned training examples.

\subsubsection{Quantitative Analysis of Misalignment Issue}
We next conduct a quantitative evaluation of these models, using three common metrics (PSNR, SSIM~\cite{wang2004image}, and LPIPS~\cite{zhang2018unreasonable}). As shown in Table.~\ref{tab:warp}, mip-NeRF 360 with bigger MLPs (4096 channels) is consistently improved by better-aligned training data, with its PSNR score increasing by a large margin (+\textbf{0.62dB} on the warped test set). When it comes to the smaller model (1024 channels), the iterative alignment strategy still brings some improvement (+\textbf{0.23dB} PSNR), although it is smaller due to underparameterization.
% architecture is not able to produce clear rendered images and that further hurts the optical flow estimation.

Although this alignment strategy produces good results, it is very time-consuming, as it requires retraining mip-NeRF 360 from scratch and re-rendering the entire training dataset multiple times. In contrast, the proposed solution in Section~\ref{sec:method} improves the baseline by +\textbf{0.23dB} PSNR on the warped test set, with little additional cost.
% , both two plans takes a huge time to make improvement, due to the wasteful process of rendering and retraining in the iterative refinement data engine. In contrast to them, the proposed methods improves the baseline with little additional cost.
% Meanwhile, both two plans takes a huge time to make improvement, due to the wasteful process of rendering and retraining in the iterative refinement data engine. In contrast to them, the proposed methods improves the baseline with little additional cost.
% As shown in the last column of Fig.~\ref{fig:flow_iteration}, mip-NeRF 360 with iterative refinement data engine successfully produce visually pleasing images and largely outperforms other baselines. We illustrate more training details in Sec.~\ref{sec:training}

\subsection{Comparing with Previous Methods}
We first evaluate our method and previous works on the high-resolution ($2560\times 1680$) ``outdoor'' scenes collected by ~\cite{mipnerf360}. For a fair comparison, we apply the proposed AligNeRF techniques to mip-NeRF 360, and take care to not increase training time with our staged training (pre-training + fine-tuning). By default, mip-NeRF 360 is trained for 250k iterations. However, since this experiment uses higher resolution images, we also look at results where we increase the training time by $4\times$ to keep the same number of training epochs. As shown in Table~\ref{tab:high_res}, NeRF~\cite{nerf} and mip-NeRF~\cite{mipnerf} exhibit poor performance, as they are not designed for 360 degree unbounded scenes. Increasing the parameters of mip-NeRF brings a small improvement, but makes the training time longer. mip-NeRF 360~\cite{mipnerf360} serves as a strong baseline, reaching 23.88dB and 24.83dB PSNR on standard and warped test set. Our proposed method outperforms the baseline methods in both groups, without introducing significant training overhead. 
% In particular, we observe that our method can introduce a large margin (+\textbf{0.39dB} PSNR on the warped test set) when trained for longer, as our alignment strategy is more effective when it is applied to a stronger baseline. 
We also include visual examples in Fig.~\ref{fig:comparison} and supplemental material, where our method produces sharper and clearer textures than all other approaches.

In Table~\ref{tab:low_res}, we analyze how our method works on lower-resolution data by running on the ``outdoor'' scenes at the same resolution ($1280\times840$) used by~\cite{mipnerf360}. The scores for prior approaches are mostly taken directly from~\cite{mipnerf360}. 
However, we also include the recent instant-NGP~\cite{mueller2022instant} method for comparison. We reached out to the authors and tuned instant-NGP~\cite{mueller2022instant} for large scenes by increasing the size of the hash grid ($2^{21}$), training batches ($2^{20}$) and the bounding box of the scene ($32$). 
%
% Similarly to the high-resolution experiments, 
% mip-NeRF 360~\cite{mipnerf360} reaches the best score among the baselines. Meanwhile, our alignment-aware training further improves the performance of mip-NeRF 360 (+\textbf{0.19dB} PSNR), even though the misalignment issue is much less severe on low-resolution images. 
Although Stable View Synthesis (SVS)~\cite{riegler2021stable} reaches the best LPIPS score, their visual results are of lower quality than other methods, as demonstrated in~\cite{mipnerf360}. 
Among these approaches, our method demonstrates the best performance among three metrics, even though the misalignment issue is much less severe on low-resolution images.

\begin{table}[!t]
  \caption{Quantitative comparison on the low resolution version ($1280 \times 840$) of the ``outdoor'' dataset~\cite{mipnerf360}.}
  \label{tab:low_res}
  \resizebox{0.99\linewidth}{!}{
  \begin{tabular}{l|ccc|c|c}
\toprule
    Methods &PSNR $\uparrow$&SSIM $\uparrow$&LPIPS $\downarrow$  &Time (hrs)& \#Params \\ 
    \hline
    NeRF~\cite{nerf} & 21.46 & 0.458 & 0.515 & 4.16 & 1.5M\\
    % NeRF w/ DONeRF param. & 21.90 & 0.472 & 0.511 & 4.59 & 1.4M\\
    mip-NeRF~\cite{mipnerf} & 21.69 & 0.471 & 0.505 & 3.17 & 0.7M\\
    NeRF++~\cite{nerfpp}& 22.76 & 0.548 & 0.427 & 9.45 & 2.4M\\ 
    Deep Blending~\cite{hedman2018deep} & 21.54 & 0.524 & 0.364 & - & -\\ 
    Point-Based~\cite{kopanas2021point} & 21.66 & 0.612 & 0.302 & - & -\\ 
    Instant-NGP~\cite{mueller2022instant} & 22.90 & 0.566 & 0.371 & 0.17 & 51.8M\\
    Stable View Synthesis~\cite{riegler2021stable} & 23.01 & \cellcolor{yellow!25}0.662 & \cellcolor{red!25}0.253 & - & - \\ 
    mip-NeRF~\cite{mipnerf} w/bigger MLPs & 22.98 & 0.625 & 0.348 & 22.71 & 9.0M\\ 
    NeRF++~\cite{nerfpp} w/bigger MLPs & \cellcolor{yellow!25}23.80 & 0.642 & 0.338 & 19.88 & 9.0M\\ 
    mip-NeRF 360~\cite{mipnerf360} & \cellcolor{orange!25}24.36 & \cellcolor{orange!25}0.689 & \cellcolor{yellow!25}0.280 & 6.89 & 9.9M\\ 
    \hline
    Ours & \cellcolor{red!25}24.55& \cellcolor{red!25}0.703& \cellcolor{orange!25}0.263& 7.12 & 10.4\\
    \bottomrule
\end{tabular}
}
\end{table}

\subsection{Ablation Study}
In Table~\ref{tab:ablation}, we conduct an ablation study of our method on the ``\textit{bicycle}'' scene (with $2560\times1680$ resolution) from the ``outdoor'' dataset~\cite{mipnerf360}. We first train a baseline mip-NeRF 360 model for 1000k iterations.
% This achieves 24.17dB and 24.96dB PSNR on the standard and warped test set, respectively.
Next, we train another mip-NeRF 360 model for 600k iterations, and apply our convolution-augmented architecture on top of it. We further fine-tune this new architecture with 200k iterations, using our patch-wise sampling strategy. This makes the convolution augmented model reach much higher quality (+\textbf{0.24dB}/\textbf{0.24dB} on standard/warped test set), without significantly increasing the training time. Meanwhile, if we additionally add the alignment-aware loss together to this convolution-augmented architecture, it further improves quality.
% , reaching 24.51dB and 25.45dB PSNR on standard and warped test set, respectively.
In particular, we observe that the improvement on warped test set (+\textbf{0.25dB} PSNR) is larger than standard test set (+\textbf{0.10dB} PSNR) after applying the proposed alignment-aware training, demonstrating that it bring more improvement in misaligned parts of the scene. Finally, the proposed high-frequency loss improves the latest results by a small margin. In contrast to other previous perceptual losses~\cite{johnson2016perceptual} which tend to improve the LPIPS score at the expense of other metrics, our loss improves all three metrics.

\begin{table}[!t]
  \caption{Ablation study for the proposed components. All results are reported on the ``\textit{bicycle}'' scene ($2560 \times 1680$) from the ``outdoor'' dataset~\cite{mipnerf360} with $4\times$ longer training. We report metrics on both the \underline{standard} and \underline{warped} test sets respectively, split by ``/''.}
    \label{tab:ablation}
    \resizebox{0.99\linewidth}{!}{
    \begin{tabular}{l|ccc|c|c}
    \toprule
    Methods &PSNR $\uparrow$&SSIM $\uparrow$&LPIPS $\downarrow$ & Time & \#Params \\ 
    \hline
    Baseline & 24.17/24.96 & 0.669/0.715 & 0.348/0.333 & 27.56 & 9.9M\\
    + Convolution &\cellcolor{yellow!25}24.41/25.20&\cellcolor{yellow!25}0.679/0.724 &\cellcolor{yellow!25}0.339/0.317 & 28.2& 10.4M\\
    + Alignment &\cellcolor{orange!25}24.51/25.45&\cellcolor{orange!25}0.679/0.729&\cellcolor{orange!25}0.337/0.312& 28.40& 10.4M\\
    + High-frequency loss &\cellcolor{red!25}24.55/25.51 &\cellcolor{red!25}0.683/0.734&\cellcolor{red!25}0.331/0.306 & 28.48& 10.4M \\
    \bottomrule
\end{tabular}
}
\end{table}

\begin{figure}[h]
    \centering
    \includegraphics[width=0.8\columnwidth]{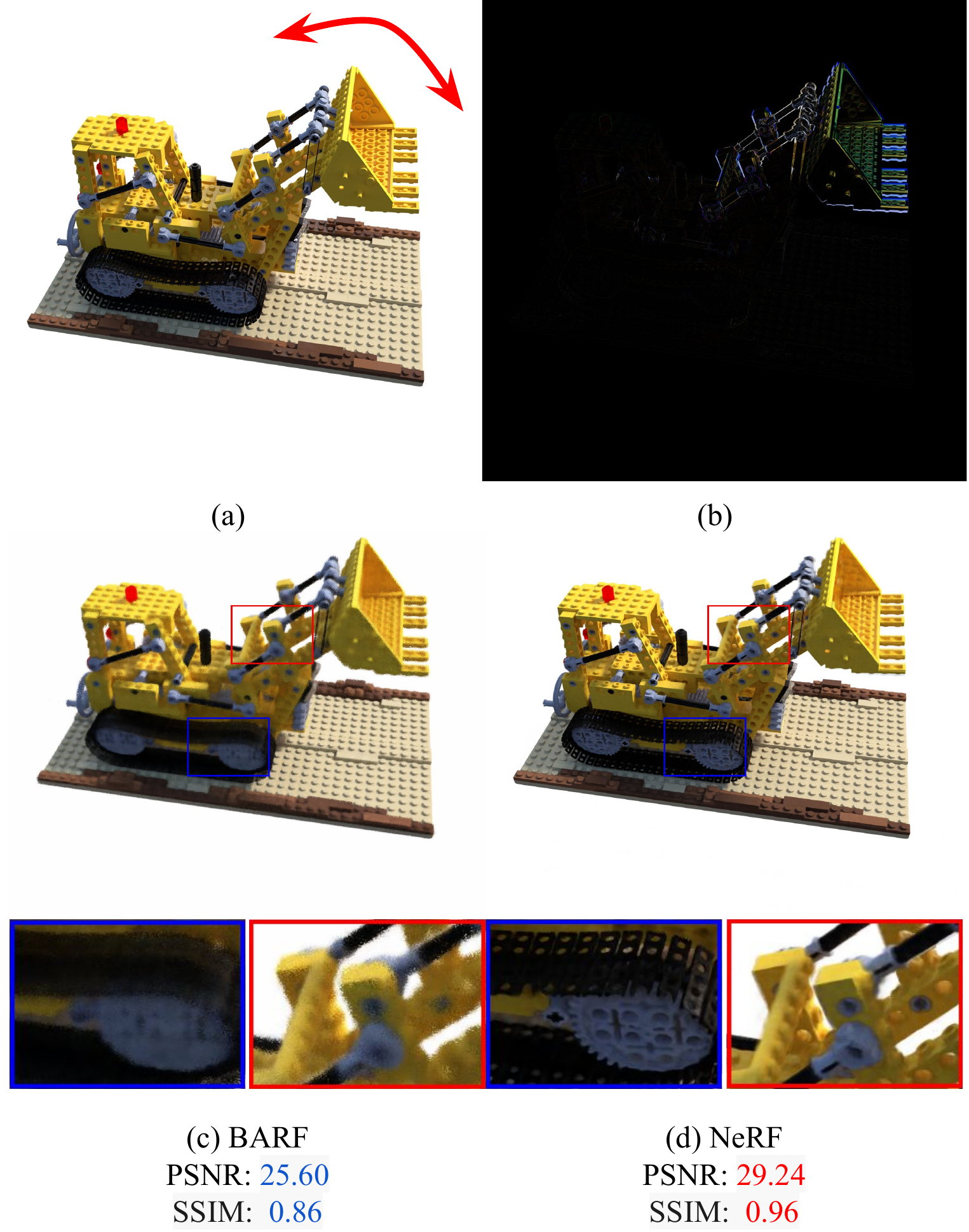}
    \caption{Visual comparison of BARF~\cite{barf} and NeRF~\cite{nerf} on moving scene of lego. (a) Illustration of ``\textit{lego}'' with moving arm. (b) Visualization of the subtle motion, calculated by the \textbf{difference} of views with two conditions. (c) Visual examples rendered by BARF. (d) Visual examples rendered by NeRF. }
    \label{fig:lego_shifted}
\end{figure}

\subsection{Evaluating Pose-free NeRFs on Non-still Scenes}
\label{sec:simulation}
One may wonder if the nonalignment issue can be fixed by jointly optimizing camera poses, e.g., using bundle-adjustment NeRF~\cite{barf}. While this strategy indeed help the static scenes, we argue that the the subtle movement of objects may even hurt its performance, as their pose optimization requires object to be static for feature matching. To demonstrate it,
we conduct a toy experiment to measure the performance of pose-free NeRF-like~\cite{barf} models on non-still scenes when subtle movement is included. 

To simulate this phenomenon, we adopt Blender software to render a new set of ``\textit{lego}'' examples with 100 training views and 200 testing views, where its arm is randomly set to be two different condition with 0.5 probability. Thus we are able to manually create misalignment. We show the visual examples in the first row of Fig.~\ref{fig:lego_shifted}, where the difference of two conditions indicates the misaligned region. We render a higher resolution ($1200 \times 1200$) to fit our high-fidelity setting.

We evaluate BARF~\cite{barf} model on this new dataset, where its initial poses are perturbed from ground truth poses. Meanwhile, we train the vanilla NeRF~\cite{nerf} for comparison. As shown in Fig.~\ref{fig:lego_shifted}, NeRF produce sharp details on the aligned region and blurry textures on the misaligned region, where BARF renders blurry results on both two regions. 
% We also include the average PSNR and SSIM on the testing set, where BARF largely behinds vanilla NeRF. 
The quantitative and qualitative experiments demonstrate that jointly optimizing NeRF and camera poses may face severe issue when moving objects are included.

%\vspace{-1.em}
\section{Conclusion}
In this work, we conducted a pilot study on training neural radiance fields on high-resolution data. We presented AligNeRF, an effective alignment-aware training strategy that improves NeRF's performance.
% AligNeRF consists of a convolution-augmented architecture that can encode more neighborhood information, an alignment-aware loss with a patch-wise matching strategy which is robust to misalignment, and an additional high-frequency aware loss to enrich texture details.
Additionally, we also quantitatively and qualitatively analyze the performance degradation brought by misaligned data, by re-generating aligned data using motion estimation. This analysis further help us to understand the current bottleneck of scaling NeRF to higher resolutions. 
% As a result, AligNeRF outperforms the current best approach without significantly increasing training costs. Nevertheless, 
we still observe that NeRF can be further improved by vastly increasing the number of parameters and by further increasing the training time.
We will investigate how to close this gap in the future.

\section{Acknowledgement}
We thank Deqing Sun and Charles Herrmann for providing codebase and pretrained model for PWC-Net, as well as constructive discussions.
% \bibliographystyle{ACM-Reference-Format}
{\small
\bibliographystyle{ieee_fullname}
\bibliography{sample-base}
}
\appendix

\end{document}